\documentclass[wcp]{jmlr}

\usepackage{amsmath,amsfonts,bm}

\def\eqref#1{equation~\ref{#1}}
\def\1{\bm{1}}

\def\vmu{{\bm{\mu}}}
\def\vtheta{{\bm{\theta}}}

\def\vl{{\bm{l}}}

\def\vr{{\bm{r}}}

\def\vw{{\bm{w}}}
\def\vx{{\bm{x}}}

\DeclareMathAlphabet{\mathsfit}{\encodingdefault}{\sfdefault}{m}{sl}
\SetMathAlphabet{\mathsfit}{bold}{\encodingdefault}{\sfdefault}{bx}{n}

\newcommand{\E}{\mathbb{E}}

\newcommand{\KL}{D_{\mathrm{KL}}}

\usepackage{xcolor}         % colors
\usepackage{algorithmic}

\usepackage{hhline}
\usepackage{multirow}
\usepackage{wrapfig}
\usepackage{caption}
\usepackage{diagbox}

\usepackage{longtable}% for long tables

\usepackage{booktabs}
\definecolor{mygray}{gray}{0.6}

\newcommand{\vect}[1]{\boldsymbol{#1}}
\newcommand{\RN}[1]{%
	\textup{\lowercase\expandafter{\it \romannumeral#1}}%
}

\makeatletter
\let\Ginclude@graphics\@org@Ginclude@graphics
\makeatother

\jmlrvolume{189}
\jmlryear{2022}
\jmlrworkshop{ACML 2022}

\title[BayesAdapter]{BayesAdapter: Being Bayesian, Inexpensively and Reliably, via Bayesian Fine-tuning}

  \author{\Name{Zhijie Deng} \Email{zhijied@sjtu.edu.cn}\\
  \addr Qing Yuan Research Institute, Shanghai Jiao Tong University
  \AND
  \Name{Jun Zhu} \Email{dcszj@mail.tsinghua.edu.cn}\\
  \addr Dept. of Comp. Sci. \& Tech., BNRist Center, THU-Bosch Joint ML Center, Tsinghua University
 }

\editors{Emtiyaz Khan and Mehmet G\"{o}nen}

\begin{document}

\maketitle

\begin{abstract}
Despite their theoretical appealingness, Bayesian neural networks (BNNs) are left behind in real-world adoption, mainly due to persistent concerns on their scalability, accessibility, and reliability. In this work, we develop the \emph{BayesAdapter} framework to relieve these concerns. In particular, we propose to adapt pre-trained deterministic NNs to be variational BNNs via cost-effective \emph{Bayesian fine-tuning}. Technically, we develop a modularized implementation for the learning of variational BNNs, and refurbish the generally applicable \emph{exemplar reparameterization} trick through exemplar parallelization to efficiently reduce the gradient variance in stochastic variational inference. Based on the lightweight Bayesian learning paradigm, we conduct extensive experiments on a variety of benchmarks, and show that our method can consistently induce posteriors with higher quality than competitive baselines, yet significantly reducing training overheads. Code is available at \url{https://github.com/thudzj/ScalableBDL}.
\end{abstract}

% \vspace{-1.5ex}
\section{Introduction}
% \vspace{-.5ex}
Much effort has been devoted to %It has been chased by the community for a long time to
developing expressive Bayesian neural networks (BNNs) to make accurate and reliable decisions~\citep{mackay1992practical,neal1995bayesian,graves2011practical,blundell2015weight}.
The principled uncertainty quantification capacity of BNNs %derived under Bayes theorem
is critical for realistic decision-making, finding applications in scenarios ranging from model-based reinforcement learning~\citep{depeweg2016learning},
active learning~\citep{hernandez2015probabilistic}
to healthcare~\citep{leibig2017leveraging} and autonomous driving~\citep{kendall2017uncertainties}.
BNNs are also known to be capable of resisting over-fitting and over-confidence.

Nonetheless, BNNs are falling far behind in terms of adoption in real-world applications compared with deterministic NNs~\citep{he2016deep,vaswani2017attention}, due to various issues.
For example, typical approximate inference methods for BNNs are often difficult to simultaneously maintain %have difficulties to conjoin
efficacy and scalability~\citep{zhang2018noisy,maddox2019simple}.
% the scalability of BNNs is typically limited by the difficulties of learning a complex, non-degenerate distribution from scratch in the high-dimensional parameter space containing many symmetries~\citep{liu2016stein,louizos2017multiplicative,sun2018functional}.
Implementing a BNN algorithm requires substantially more expertise than implementing a deterministic NN program.
% BNNs' need of more expertise for implementation further impedes their all-round utilization.
Moreover, as revealed, BNNs trained from scratch without {the} ``cold posterior'' trick are often systematically worse than their point-estimate counterparts in terms of predictive performance~\citep{wenzel2020good}; some easy-to-use BNNs (e.g., Monte Carlo dropout) tend to suffer from mode collapse in function space, thus usually give uncertainty estimates of poor fidelity~\citep{fort2019deep}.

% BNNs, as a special class of neural networks, are still black-box models to handle high-dim. data, so their inherent high nonlinearity and rare interpretability may render the uncertainty quantification acquired from Bayes' rule unreliable in practice.

% Though the reliability of BNNs' uncertainty quantification is theoretically guaranteed by Bayes' rule, the black-box nature of neural networks and the high-dim. nature of input data are likely to ruin the self-guaranteed uncertainty estimation in practice.
% the existing BNNs can hardly distinguish normal data from practically challenging data, e.g., adversarial and fake samples, due to the black-box nature of neural networks.
% the uncertainty outcomes of the existing BNNs are typically unreliable in industrial scenarios for malicious input like adversarial or fake samples if the model have not seen them during learning.
% , the BNNs are expected to deliver reliable predictive uncertainty for challenging input like adversarial or fake samples, which, yet, cannot be guaranteed due to the unsupervised manner of the acquisition of uncertainty quantification. %, the black-box nature of deep networks, and the high-dimensional nature of input data.
% These obstacles are exacerbated when pushing the modeling limit to larger datasets and % (e.g., ImageNet~\citep{imagenet_cvpr09}),
% deeper architectures. % (e.g., ResNets~\citep{he2016deep}),
% and realistic tasks.% (e.g., face recognition~\citep{deng2019arcface}).

\begin{figure}[t]
% \vspace{-0.3ex}
     \centering
    %  \begin{subfigure}[b]
        %  \centering
         \includegraphics[width=0.52\linewidth]{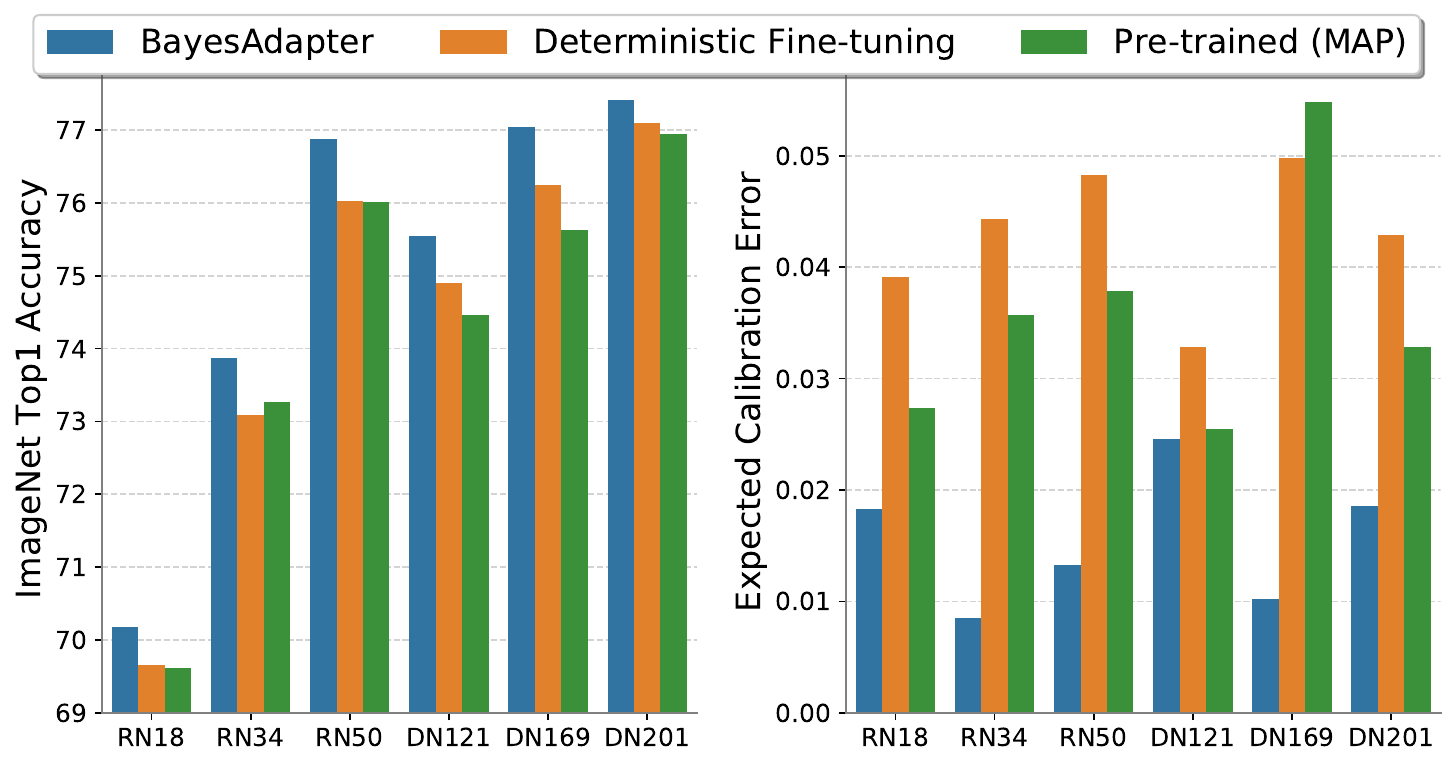}
        % \vspace{-2.5ex}
        % \caption{\scriptsize Comparison on accuracy and model calibration. (ImageNet)}
        % \label{fig:mode_col}
    %  \end{subfigure}
    %  \hfill
    %  \begin{subfigure}[b]
        %  \centering
        % \vspace{3ex}
        \includegraphics[width=0.47\linewidth]{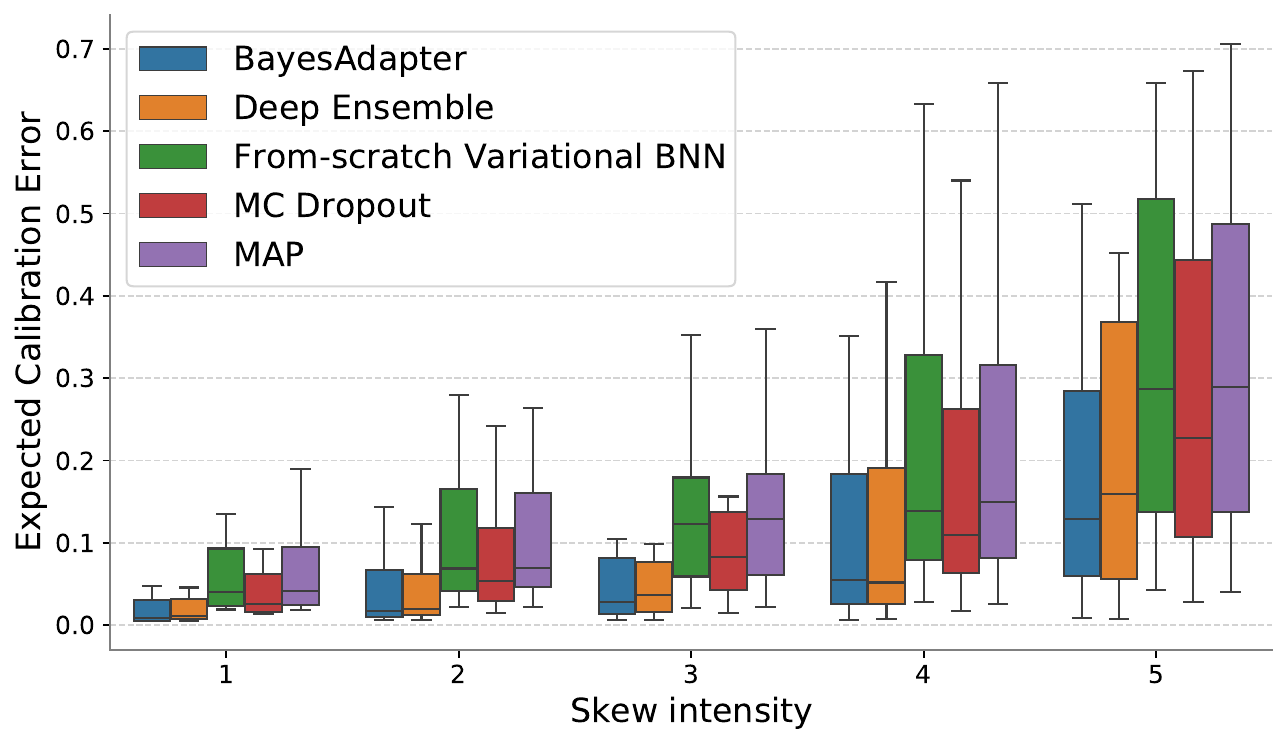}
        % \vspace{-2.5ex}
        % \caption{\scriptsize Comparison on model calibration for CIFAR-10 corruptions.}
        % \label{fig:rej-dec}
    %  \end{subfigure}
    %  \vspace{-1.5ex}
     \caption{\small (Left): BayesAdapter boosts the accuracy of ImageNet classifiers without compromising model calibration (estimated by expected calibration error (ECE)~\citep{guo2017calibration}).
     By contrast, deterministic fine-tuning only marginally improves the accuracy of pre-trained models yet aggravates over-confidence.
    %  fails in both axes\junz{say "fail" is too rude: it does improve accuracy in most cases over MAP; treat the competitors more fairly, to avoid offending reviewers. Also, say "axes" is confusing, reviewers may look at the axis of each plot...}.
     RN refers to ResNet~\citep{he2016deep} and DN refers to DenseNet~\citep{huang2017densely}.
     (Right): BayesAdapter learns a CIFAR-10 classifier which approaches or outperforms competing baselines in terms of ECE for CIFAR-10 corruptions~\citep{hendrycks2019benchmarking}. Each box summarizes the ECE across 19 types of skew.
     We perform \emph{Bayesian fine-tuning} for only \textbf{4} and \textbf{12} epochs on ImageNet and CIFAR-10 respectively. More details are deferred to Sec~\ref{sec:exp}.}
     \label{fig:acc-comp}
    %  \vspace{-1ex}
\end{figure}

To mitigate these issues, we develop a pre-training \& fine-tuning workflow for learning variational BNNs given an inherent connection between variational BNNs~\citep{blundell2015weight} and regular deep neural networks (DNNs).
The resultant \emph{BayesAdapter} framework learns a variational BNN by performing several rounds of \emph{Bayesian fine-tuning}, starting from a pre-trained deterministic NN.
BayesAdapter is effective and lightweight, and conjoins the complementary benefits from deterministic training and Bayesian reasoning, e.g., performance matching the pre-trained deterministic models, resistance to over-fitting, reliable uncertainty estimates, etc. (find evidence in Figure~\ref{fig:acc-comp}).
% The converged parameters of the deterministic model serve as a strong start point for \emph{Bayesian fine-tuning}, allowing to discover qualified function modes more easily than the from-scratch Bayesian learning.
%(verified by Sec~\ref{sec:perf}).

% BayesAdapter learns accurate image classifiers with calibrated uncertainty at a low cost

To improve the usability of \emph{BayesAdapter}, we provide a modularized implementation for the stochastic variational inference (SVI) under multiple representative variational distributions, including \emph{mean-field Gaussian} and \emph{parameter-sharing ensemble}.
Reducing the variance of stochastic gradients is crucial for stabilizing and accelerating SVI, while the pioneering works such as local reparameterization~\citep{kingma2015variational} and Flipout~\citep{wen2018flipout} can only deal with specific variational distributions, e.g., Gaussians and distributions whose samples can be reparameterized with symmetric perturbations.
% ~\citep{kingma2015variational,wen2018flipout} can only handle some variationals with restrictive forms\junz{this claim is not concrete, thus not convincing, again may offend reviewers. Make it precise, e.g., restrictive in what? What problem exactly are we facing. When you claim the weakness of existing work, should make it very precise and concrete.}.
To tackle this issue, we refurbish the widely-criticized \emph{exemplar reparameterization}~\citep{kingma2015variational} by accelerating the exemplar-wise computations through parallelization, giving rise to an efficient and general-purpose gradient variance reduction technique.
% To stabilize the stochastic variational inference beyond Gaussian variational or, we
% devise a general-purpose gradient variance reduction strategy named \emph{exemplar reparametrization}. % and implement it as a native feature of the library.
% {Based on} these designs, we perform an investigation on \emph{do variational BNNs know what they do not know}, and find that typical variational BNNs can seldom yield calibrated uncertainty estimates on realistic, malicious out-of-distribution (OOD) data.
% We then provide a corresponding prescription, where Bayesian inference is augmented with biased yet meaningful regularization, to ameliorate this pathology.
% As a solution, we opt to explicitly regularize the variational BNNs to behave uncertainly on a collection of OOD data during \emph{Bayesian fine-tuning}.
% This regularization takes the form of a margin loss, and
% This loss rectifies the uncertainty estimation according to highly customizable, low-cost uncertainty supervisions and
% is readily applicable to most of the existing BNNs.
% Given such design, we can promise that the uncertainty estimates are reliable at least on the forseeable, challenging instances.
% Figure~\ref{fig:framework} depicts our whole framework.

We conduct extensive experiments to validate the advantages of \emph{BayesAdapter} over competing baselines, in aspects covering efficiency, predictive performance, and quality of uncertainty estimates.
Desirably, we scale up \emph{BayesAdapter} to big data (e.g., ImageNet~\citep{imagenet_cvpr09}), deep architectures (e.g., ResNets~\citep{he2016deep}), and practical scenarios (e.g., face recognition~\citep{deng2019arcface}), and observe promising results.
We also perform a series of ablation studies to reveal the characteristics of the proposed approach.

\iffalse
In summary, our contributions are as follows:
\begin{enumerate}
    \vspace{-0.1cm}
    \item We propose \emph{BayesAdapter}, to quickly and cheaply adapt a pre-trained DNN to be Bayesian without compromising performance when facing new tasks. %a two-step framework composed of \emph{deterministic pre-training} and \emph{Bayesian fine-tuning}, to obtain practical BNNs.
    \vspace{-0.03cm}
    \item We provide an easy-to-use instantiation of stochastic VI, which allows learning a BNN as if training a deterministic NN and frees the users from tedious details of BNN.
    \vspace{-0.03cm}
    \item We augment the fine-tuning with a generally applicable uncertainty regularization term to rectify the predictive uncertainty according to a collection of OOD data.
    \vspace{-0.03cm}
    \item Extensive studies validate that BayesAdapter is scalable; the delivered BNN models are high-quality; and the acquired uncertainty quantification is calibrated and transferable.
\end{enumerate}
\fi

% Figure~\ref{fig:framework} presents the procedure of BayesAdapter.
% The above adaptation procedure is model agnostic and thus is readily applicable to a wide spectrum of deep models to deal with data in various modality.

% \vspace{-.5ex}
\section{Background}
% \vspace{-.5ex}

In this section, we motivate \emph{BayesAdapter} by drawing a connection between variational BNNs and DNNs trained by \emph{maximum a posteriori} (MAP) estimation.

%a practical and robust implementation of stochastic variational inference for \emph{Bayesian fine-tuning}.

% BayesAdapter by answering three core and coherent questions: $(\RN{1})$ \emph{How to convert a DNN to be a BNN without the loss of the efforts for pre-training?} $(\RN{2})$ \emph{How to efficiently fine-tune the resultant BNN with variance reduced gradients?} $(\RN{3})$ \emph{How to augment the training with uncertainty supervision to pursue more calibrated uncertainty estimates?} After all, we briefly describe the overall algorithmic procedure of BayesAdapter.

% \vspace{-0.2cm}

% \vspace{-.5ex}
% \subsection{Background}
% \vspace{-.5ex}
Let $\mathcal{D}=\{(\vx^{(i)}, y^{(i)})\}_{i=1}^n$ be a given training set, where $\vx^{(i)} \in \mathbb{R}^d$ and $y^{(i)} \in \mathcal{Y}$ denote the input data and label, respectively.
A DNN model can be fit via MAP estimation:
\begin{equation}
\small
\label{eq:map}
    {\max_{\vect{w}} \frac{1}{n}\sum_{i} [\log p(y^{(i)}| \vx^{(i)}; \vect{w})] + \frac{1}{n}\log p(\vect{w})}.
% \vspace{-0.05cm}
\end{equation}
We use $\vect{w} \in \mathbb{R}^p$ to denote the high-dimensional model parameters, with $p(y| \vect{x}; \vect{w})$ as the predictive distribution associated with the model.
The prior $p(\vect{w})$, when taking the form of an isotropic Gaussian $\mathcal{N}(\vect{w}; \vect{0}, \sigma^2_0\mathbf{I})$, reduces to the weight decay regularizer with coefficient $\lambda=1/(\sigma_0^2n)$ in optimization.
Nevertheless, deterministic training may easily cause over-fitting and over-confidence, rendering the learned models of poor reliability (see Figure~\ref{fig:acc-comp}).
% More critically, the predictive confidence outputted by DNNs has low fidelity, unable to convey enough information to us for deciding whether the predictions are trustable or not.
% All these issues together raise the requirement for more principled deep models with better statistical and mathematical properties, with the BNN as a promising candidate.
Naturally, BNNs come into the picture to address these limitations.

Typically, BNNs learn by inferring the posterior $p(\vect{w}|\mathcal{D})$ given the prior $p(\vect{w})$ and the likelihood $p(\mathcal{D}|\vect{w})$.
Among the wide spectrum of BNN algorithms~\citep{mackay1992practical,neal1995bayesian,graves2011practical,blundell2015weight,liu2016stein,gal2016dropout}, variational BNNs are particularly promising due to their analogy to ordinary backprop. % one, rendering the learning in the style of training DNNs.
Formally, variational BNNs use a $\vect{\theta}$-parameterized variational distribution $q(\vect{w}|\vect{\theta})$ to approximate $p(\vect{w}|\mathcal{D})$, by maximizing the evidence lower bound (ELBO) (scaled by $1/n$):
\begin{equation}
\small
\label{eq:elbo}
% \vspace{-0.1cm}
    {\max_{\vect{\theta}} \underbrace{\E_{q(\vect{w}|\vect{\theta})} \big[\frac{1}{n}\sum_{i} \log p(y^{(i)}|\vx^{(i)};\vect{w})\big]}_{\mathcal{L}_{ell}} \underbrace{\vphantom{\big[\frac{1}{n}\sum_{i} \log p(y^{(i)}|\vx^{(i)};\vect{w})\big]}- \frac{1}{n} \KL\left(q(\vect{w}|\vect{\theta})\Vert p(\vect{w})\right)}_{\mathcal{L}_c}},
% \vspace{-0.05cm}
\end{equation}
where $\mathcal{L}_{ell}$ is the \emph{expected log-likelihood} and $\mathcal{L}_c$ is the \emph{complexity loss}.
By casting posterior inference into optimization, Eq.~(\ref{eq:elbo}) makes the training of BNNs resemble that of DNNs.
After training, the variational posterior is leveraged for prediction through marginalization:
\begin{equation}
% \vspace{-0.05cm}
\label{eq:pred}
\small
    p(y|\vect{x}, \mathcal{D}) \approx \E_{q(\vect{w}|\vect{\theta})} p(y|\vect{x}; \vect{w}) \approx \frac{1}{S}\sum_{s=1}^S p(y|\vect{x}; \vect{w}^{(s)}),
% \vspace{-0.05cm}
\end{equation}
{where} $\vect{w}^{(s)} \sim  q(\vect{w}|\vect{\theta}), s=1,...,S$, with $S$ {denoting} the number of Monte Carlo (MC) samples.
Eq.~(\ref{eq:pred}) is known as \emph{posterior predictive}, \emph{Bayes ensemble}, or \emph{Bayes  model average}.

We can simultaneously quantify the \emph{epistemic} uncertainty with these MC samples.
A principled uncertainty metric is the mutual information between the model parameter and the prediction~\citep{smith2018understanding}, estimated by ($H$ denotes the Shannon entropy):
\begin{equation}
% \vspace{0.15cm}
\small
\label{eq:mi}
    \mathcal{I}(\vw, y|\vx, \mathcal{D}) \approx H\left(\frac{1}{S}\sum_{s=1}^S p(y|\vect{x}; \vect{w}^{(s)})\right) - \frac{1}{S}\sum_{s=1}^S H\left(p(y|\vect{x}; \vect{w}^{(s)})\right).
\end{equation}
% where .

However, most of the existing variational BNNs exhibit limitations in scalability and performance~\citep{osawa2019practical,wenzel2020good}, compared with their deterministic counterparts.
This is mainly due to the higher difficulty of learning high-dimensional distributions from scratch than point estimates. %, and challenges in finding non-degenerated optima of  highly nonlinear functions characterized by NNs.
%and that the posterior may be stuck in the extensive degenerate solutions, stemming from the high nonlinearity of neural networks.

Given that MAP converges to a \emph{mode} of the Bayesian posterior, it might be plausible to \emph{adapt pre-trained deterministic DNNs to be Bayesian economically}. Following this hypothesis, we repurpose the converged parameters $\vect{w}^*$ of MAP -- take $\vect{w}^*$ as the initialization of the parameters of the approximate posterior.
%as a Gaussian $\mathcal{N}(\vect{w}; \vtheta)$ with $\vtheta=(\vmu, \vect{\Sigma})$, where $\vect{\mu}$ is initialized as $\vect{w}^*$ and $\vect{\Sigma} \in \mathbb{R}^{p\times p}$ denotes the covariance.
% Then, we arrive at a BNN with \emph{posterior predictive}:
% , where $\vmu$ is perturbed, and the predictions from multiple likely models are assembled.
% Intuitively, the BNN model structurally perturbs $\vmu$ to obtain multiple likely models and then assembles their predictions.
% $\vect{\Sigma}$ controls the magnitude of perturbation.
Laplace approximation~\citep{mackay1992practical} is a classic method in this spirit, which assumes a Gaussian approximate posterior, and {adapts} $\vw^*$ and the local curvature at $\vw^*$ as the Gaussian mean and variance respectively.
% to generate an informative $\vect{\Sigma}$ is by ,
%\junz{more elaboration on Laplace to better support the argument on its weakness (e.g., the so-claimed "imperfect-ness", and motivate fine-tuning.} %its inflexibility hinders the joint adaptation of the approximate posterior w.r.t. the data, and
Yet, Laplace approximation is inflexible and usually computationally prohibitive (only after the introduction of Gauss-Newton approximation, KFAC approximation, and particularly the last-layer approximation, the cost of Laplace approximation becomes affordable~\citep{daxberger2021laplace}). %has two limitations: 1) it is more like a postprocessing, lacking the flexibility to jointly adapt the mean and covariance of the Gaussian posterior w.r.t. data; 2) its naive implementation without strong assumptions may be computationally prohibitive.
% the involved Hessian calculation may be hard to implement in and matrix inversion, and is computationally prohibitive.
Alternatively, we develop the more practical \emph{Bayesian fine-tuning} scheme, whose core notion is to fine-tune the imperfect approximate posterior by maximizing ELBO.
% Algorithm~\ref{algo:1} gives an overview of \emph{Bayesian fine-tuning}.
%advocating fine-tuning the approximate posterior following the stochastic VI pipeline and meanwhile providing two more accessible variational configurations.

% Instead, we suggest a more practical workflow -- that fine-tunes the approximate posterior $\mathcal{N}(\vect{w}; \vect{\mu}, \vect{\Sigma})$ by maximizing the ELBO with randomly initialized $\vect{\Sigma}$.

\begin{figure}[t]
% \vspace{-2ex}
    \centering
    \includegraphics[width=0.55\linewidth]{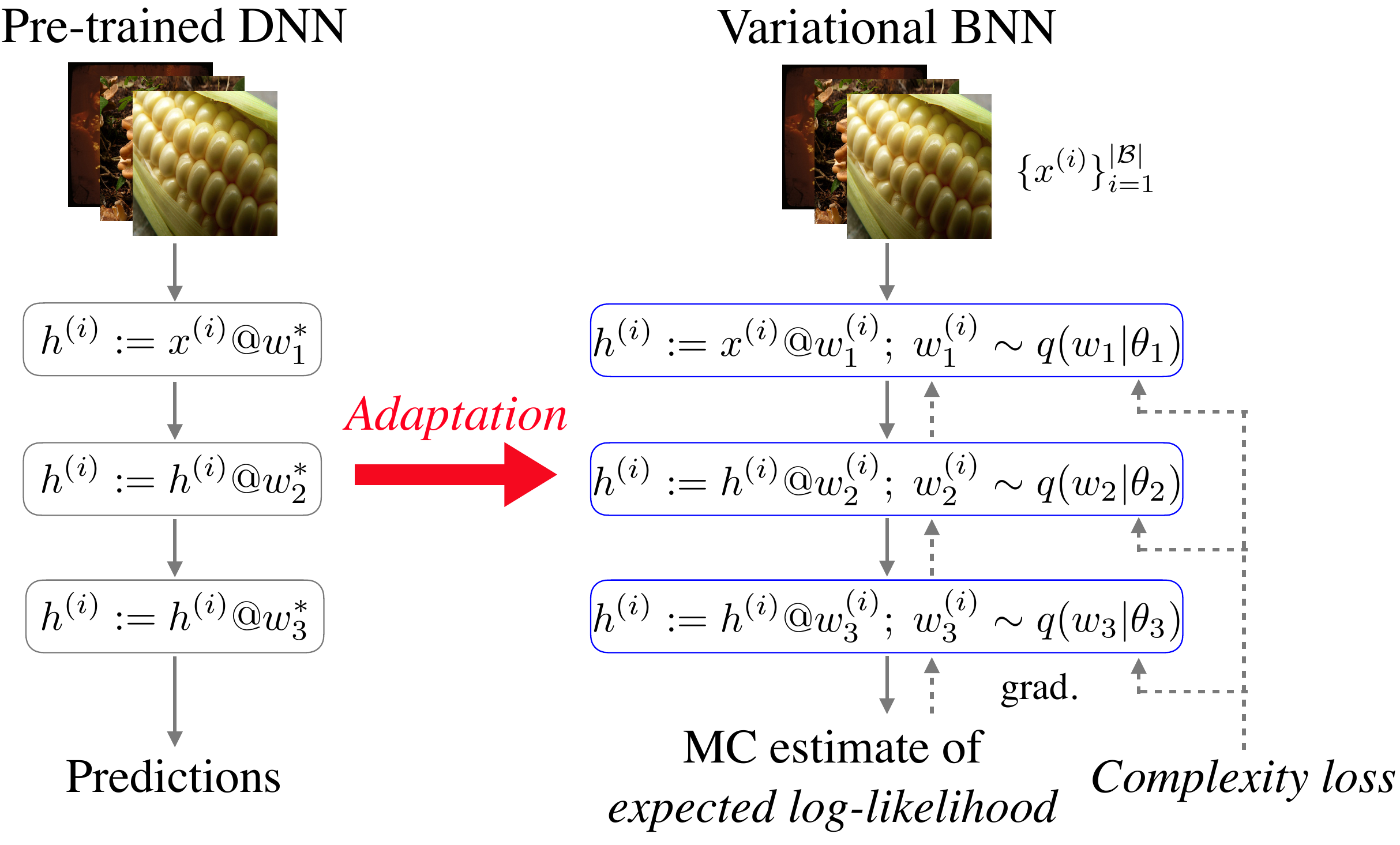}
    % \vspace{-1.3ex}
    % \label{fig:framework}
\caption{\small \emph{BayesAdapter} adapts pre-trained DNNs to be variational BNNs and then launches few rounds of \emph{Bayesian fine-tuning}.
We provide a modularized implementation for \emph{Bayesian fine-tuning}, allowing users to learn a variational BNN as if training a DNN under a weight decay regularizer. %\junz{move to next page}
% which directly sends the gradients of the \emph{complexity loss} back to the variantional parameters like weight decay, and only explicitly deals with the \emph{expected log-likelihood} via Monte Carlo estimation.
% To reduce the variance of stochastic gradients, we further develop the \emph{exemplar reparameterization} technique, which deploys a dedicated parameter sample for every exemplar in the mini-batch.
}
\label{fig:framework}
% \vspace{-10.ex}
\end{figure}

% \vspace{-0.1cm}
\section{BayesAdapter}
% \vspace{-0.1cm}
\label{sec:ft}

We describe \emph{BayesAdapter} in this section.  Figure~\ref{fig:framework} gives its illustration.

In \emph{BayesAdapter}, the configuration of the variational distribution $q(\vw|\vtheta)$ plays a decisive role.
Although a wealth of variationals have emerged for adoption~\citep{louizos2016structured,li2017gradient,wen2018flipout},
on one hand, more complicated ones~\citep{louizos2017multiplicative,shi2018spectral} are routinely accompanied by less scalable learning; on the other hand, the aforementioned hypothesis inspiring \emph{Bayesian fine-tuning} entails an explicit alignment between the DNN parameters $\vw^*$ and the variational parameters $\vtheta$.
Thus, we primarily concern the typical \emph{mean-field Gaussian} distribution as well as a more powerful one that resembles {Deep Ensemble}~\citep{lakshminarayanan2017simple}.

% Consequently, we at first place our attention on the typical mean-field Gaussian variational.
% \junz{To help readers, better to explain here why we have MFG and PSE, and how are they related? are you suggesting both or recommend one?}

% In the following, we instantiate \emph{Bayesian fine-tuning} on two representative variational configurations and develop a unified user-friendly implementation.
% In the following, we discuss how to deal with each term in Eq.~(\ref{eq:elbo}).

% \zhijie{move to a suitable place}
% Algorithm~\ref{algo:1} gives an overview of \emph{BayesAdapter}.
% We at first specify the detailed configuration of the BNN model. %along with the development of stochastic gradient estimator used for fine-tuning.

\subsection{Mean-field Gaussian (MFG) Variational}
Without losing generality, we write the \emph{MFG} variational as $q(\vw|\vtheta)=\mathcal{N}(\vw; \vmu, \textbf{diag}(\exp(2\vect{\psi})))$, with $\vmu, \vect{\psi}\in \mathbb{R}^p$ denoting the mean and the logarithm of standard deviation respectively.
In this sense, we can naturally initialize $\vmu$ with $\vw^*$ at the beginning of fine-tuning to ease approximate inference and to enable the investigation of more qualified posterior modes.
As in MAP, we also assume an isotropic Gaussian prior $p(\vect{w})=\mathcal{N}(\vect{w}; \vect{0}, \sigma^2_0\mathbf{I})$.
% Then the complexity loss boils down to:
% \begin{equation}
% \small
% \label{eq:kl}
% \vspace{-0.1cm}
% \begin{aligned}
%     \mathcal{L}_c & = -\frac{1}{n}\KL\left(\mathcal{N}(\vw; \vmu, \textbf{diag}(\exp(2\vect{\psi}))) \Vert \mathcal{N}(\vect{w}; \vect{0}, \sigma^2_0\mathbf{I})\right) \\ %= -\frac{\vmu^T\vmu+\text{tr}(\vect{\Sigma})}{2\sigma^2_0n} + \frac{\log \text{det} \vect{\Sigma}}{2n} + c, \\
%     &=-\frac{\sum_{j}\vmu_j^2+\sum_{j}\exp(2\vect{\psi}_j)}{2\sigma^2_0n} + \frac{\sum_j \vect{\psi}_j}{n} + \text{constant},
% \end{aligned}
% \vspace{-0.0cm}
% \end{equation}
% where the subscript $j$ denotes $j$th coordinate.
Then the gradients of the \emph{complexity loss} can be derived analytically:
\begin{equation}
\small
\label{eq:grad}
% \vspace{-0.1cm}
    \nabla_{\vmu} \mathcal{L}_c = -\lambda{\vmu}, \;\; \nabla_{\vect{\psi}} \mathcal{L}_c = -\lambda{\exp(2\vect{\psi})} + \frac{1}{n}, \;\; \text{with}\; \lambda=\frac{1}{\sigma_0^2n}.
\vspace{-0.0cm}
\end{equation}
Intuitively, the above gradients for the variational parameters correspond to a variant of the vanilla weight decay in DNNs.  %deterministic model parameters as in MAP,
% with $\lambda$ also equaling to $1/(\sigma_0^2n)$.
% The above gradients remind us of the typical weight decay applied on the deterministic neural network parameters.
Having identified this, we can implement a module similar to weight decay to implicitly be responsible for the \emph{complexity loss}, leaving only the \emph{expected log-likelihood} $\mathcal{L}_{ell}$ required to be explicitly handled.
We will elaborate on the details of solving $\max\mathcal{L}_{ell}$ after presenting a more expressive variational configuration.

\subsection{Parameter-sharing Ensemble (PSE) Variational}
Despite simplicity, the \emph{MFG} variational can be limited in expressiveness %is frequently criticized for its too limited expressiveness
for capturing the multi-modal parameter posterior of over-parameterized neural networks. %\junz{looks like you want to discard MFG, but it is not; here I softened the wording.}
Empowered by the observation that Deep Ensemble~\citep{lakshminarayanan2017simple} is a compelling Bayesian marginalization mechanism in deep learning~\citep{wilson2020bayesian}, we intend to develop a low-cost ensemble-like variational for more practical Bayesian deep learning.

Specifically, we first define the variational as a uniform mixture of $C$ Gaussians: $q(\vw|\vtheta)=\frac{1}{C}\sum_{c}\mathcal{N}(\vw; \vw^{(c)}, \vect{\Sigma}^{(c)})$, where $\vect{\Sigma}^{(c)}\in\mathbb{R}^{p\times p}$ is positive-definite and its elements are independent of the dimension $p$.\footnote{We define the variational as a mixture of Gaussians instead of a mixture of deltas to ensure the variational is \emph{absolutely continuous} w.r.t. the prior. This avoids the \emph{singularity} issue in variational inference~\citep{hron2018variational}.}
In this sense, the \emph{complexity loss} boils down to the KL divergence between a mixture of Gaussians and a Gaussian, which, yet, cannot be calculated analytically in general.
Nevertheless, under the mild assumption that $\vw^{(c)} \in \mathbb{R}^p$ is normally distributed and $p$ is large enough, the KL divergence can be approximated by a weighted sum of the KL divergences between the Gaussian components and the Gaussian prior (refer to \citep{gal2015dropout} for detailed discussion and proof). Namely,
% Formally, we have
\begin{equation}
\small
\begin{aligned}
\small
\label{eq:loss-c} &-\frac{1}{n}\KL\left(\frac{1}{C}\sum_{c}\mathcal{N}(\vw; \vw^{(c)}, \vect{\Sigma}^{(c)}) \big\Vert \mathcal{N}(\vect{w}; \vect{0}, \sigma^2_0\mathbf{I})\right) \\
\approx & -\frac{1}{2\sigma_0^2nC}\sum_{c=1}^C \left(\Vert\vw^{(c)}\Vert_2^2+\text{trace}(\vect{\Sigma}^{(c)}) - \sigma_0^2\log|\vect{\Sigma}^{(c)}|\right) +\text{constant}.
\end{aligned}
\vspace{-0.0cm}
\end{equation}

Based on the observation that Bayesian model average benefits significantly more from the exploration of new modes than navigation around a local mode~\citep{wilson2020bayesian}, we assume $\vect{\Sigma}^{(c)}, c=1,...,C$, to be a constant diagonal matrix $\sigma^2\mathbf{I}$ with $\sigma^2$ approaching ${0}$.
Namely, we purely chase multi-mode exploration and leave the joint optimization of $\vect{\Sigma}^{(c)}$ and $\vw^{(c)}$ for future investigation.
Then, $q(\vw|\vtheta)$ almost amounts to a mixture of deltas (i.e., an ensemble) and with high probability we can approximate the realisation of $\vw$ by a uniform sample from $\{\vw^{(1)}, ..., \vw^{(C)}\}$.
Meanwhile, the \emph{complexity loss} approximately becomes $-\frac{\lambda}{2C}\sum_{c=1}^C \Vert\vw^{(c)}\Vert_2^2 + \text{constant}$, and we can easily implement a weight decay-like module to be responsible for its gradient.
We comment here that it may be more plausible to alternatively leverage the rigorous quasi-KL divergence~\citep{hron2018variational} for estimating the divergence between a mixture of deltas and the Gaussian prior, left as a future work.

Simulating an ensemble is far from our ultimate goal due to the required high cost.
To make the variational economical, we explore a valuable insight from recent works~\citep{wen2020batchensemble,wenzel2020hyperparameter} that the parameters of different ensemble components can be partially shared without undermining effectiveness.

% Basically, the mixture components are generated by multiplying dedicated low-rank perturbations to shared full-rank parameters.
Specifically, abusing $\vw$ to notate the parameter matrix of size $m_{\text{in}} \times m_{\text{out}}$ in a neural network layer, %\footnote{It is known that the multidimensional convolutional kernel can be converted to an equivalent parameter matrix.}
we generate $C$ components via: $\vw^{(c)}= \vl^{(c)}\vr^{(c)} \circ \bar{\vw} , c=1,...,C$, where $\bar{\vw} \in \mathbb{R}^{m_{\text{in}} \times m_{\text{out}}}$ are the shared parameters and $\vl^{(c)} \in \mathbb{R}^{m_{\text{in}}\times r}$ and $\vr^{(c)} \in \mathbb{R}^{r \times m_{\text{out}}}$ correspond to $r$-rank decomposition of some perturbations.
$\circ$ is element-wise multiplication.
The shared parameters $\bar{\vw}$ can be initialized as $\vw^*$ to ease and speedup \emph{Bayesian fine-tuning}.
When the rank $r$ is suitably small, the above design can significantly reduce the model size, and save the training effort.
Of note that the previous works~\citep{wen2020batchensemble,wenzel2020hyperparameter} confine $r$ to be 1 to permit the adoption of a specific gradient variance reduction trick.
Conversely, we loosen this constraint by using a more generally applicable variance reduction tactic, detailed below.

\iffalse
We can then formally write the \emph{PSE} variational as $q(\vw|\vtheta) = \frac{1}{C}\sum_{c=1}^C \delta(\vw - \vl^{(c)}\vr^{(c)}\circ\bar{\vw})$ (for one NN layer here), where $\delta$ denotes the Dirac delta function.
The shared parameters $\bar{\vw}$ can be initialized as $\vw^*$ to ease and speedup the \emph{Bayesian fine-tuning}.
Also assuming an  isotropic Gaussian prior, we can trivially derive the gradients of the \emph{complexity loss} $\mathcal{L}_c$:
\vspace{-0.2cm}
\begin{equation}
% \vspace{-0.15cm}
\small
\label{eq:grad-ens}
\begin{aligned}
    \nabla_{\bar{\vw}} \mathcal{L}_c &=  - \frac{\lambda}{C}  \sum_{c=1}^C (\vl^{(c)}\vr^{(c)}) \circ (\vl^{(c)}\vr^{(c)}) \circ \bar{\vw} , \\
    \nabla_{\vl^{(c)}} \mathcal{L}_c &= -\frac{\lambda}{C} \left((\vl^{(c)}\vr^{(c)}) \circ \bar{\vw}\circ\bar{\vw}\right){\left(\vr^{(c)}\right)}^{\top}, \\
    \nabla_{\vr^{(c)}} \mathcal{L}_c &= -\frac{\lambda}{C} {\left(\vl^{(c)}\right)}^{\top}\left((\vl^{(c)}\vr^{(c)}) \circ \bar{\vw}\circ\bar{\vw}\right).
% \vspace{-0.1cm}
\end{aligned}
\end{equation}
These gradients can also be absorbed by a weight decay like gradient editing operator.
We then only need to explicitly estimate $\mathcal{L}_{ell}$, rendering the proposed approach modularized.
\fi

\subsection{A Reliable Estimation of the Expected Log-likelihood $\mathcal{L}_{ell}$}
% After that, we talk about a general-purpose variance reduction technique for stochastic VI, and offer some practical strategies to make it cope with contemporary ML frameworks.
Given the high non-linearity of deep NNs and the large volume of data in real-world scenarios, we follow the stochastic variational inference (SVI) paradigm for estimating $\mathcal{L}_{ell}$.
Formally, given a mini-batch of data $\mathcal{B}=\{(\vx^{(i)}, y^{(i)})\}_{i=1}^{|\mathcal{B}|}$, we solve
% \vspace{-0.15cm}
\begin{equation}
\small
\label{eq:loss1}
    \max_{\vect{\theta}} \mathcal{L}'_{ell}= \frac{1}{|\mathcal{B}|}\sum_{i=1}^{|\mathcal{B}|} \log p(y^{(i)}|\vx^{(i)};\vect{w}),
% \vspace{-0.15cm}
\end{equation}
where $\vw$ is drawn from the \emph{MFG} or \emph{PSE} variational via reparameterization~\citep{kingma2013auto}.
The gradients w.r.t. the variational parameters %$\vmu, \vect{\psi}$ or $\bar{\vw}, \vl^{(c)}, \vr^{(c)}, c=1,...,C$
can be derived automatically with autodiff libraries, thus the training resembles that of regular DNNs.

However, gradients derived by $\mathcal{L}'_{ell}$ might exhibit high variance,
%However, as widely criticized, gradients induced by $\mathcal{L}'_{ell}$ exhibit high variance, which, briefly speaking, is
caused by sharing the sampled parameters $\vw$ across data in the mini-batch~\citep{kingma2015variational}.
Popular techniques for addressing this issue typically assume a restrictive form of variational distribution~\citep{kingma2015variational,wen2018flipout}, struggling to handle structured distributions like the proposed \emph{PSE} with $>1$ rank.
% {Local reparameterization} has been proposed to reduce the variance, but it is typically limited to exponential family and requires at least 2x forward-backward FLOPS compared to vanilla reparameterization~\citep{kingma2015variational}. %, and in practice we observe \emph{NaN} training loss when applying this trick to learning BNNs on CIFAR-10~\citep{krizhevsky2009learning} with architecture wide-ResNet-28-10~\citep{zagoruyko2016wide}\zhijie{?}.
% often results in numerical instability with working with more complex neural architectures, such as ConvNets \hao{is there any experment results supporting this last statement?}.
% For variance reduction, local reparameterization trick~\citep{kingma2015variational} has been proposed yet it demands rigid two times of complexity for forward and backward propagation.
% What's worse, we empirically found that this trick would incur numerical instability when being applied to deep convolutional networks (ConvNets).
% Flipout~\citep{wen2018flipout} is an alternative solution, but it also works under strong assumptions, e.g., the MC estimation is based on symmetric perturbations, thus cannot handle structured distributions like the proposed \emph{PSE} variational. %Besides, it is as slow as \emph{local reparameterization}.
Fortunately, there is a generally applicable strategy for reducing gradient variance in stochastic variational inference named \emph{exemplar reparametrization} (ER), which samples dedicated parameters for every exemplar in the minibatch for estimating $\mathcal{L}_{ell}$:
% In our case, for $i = 1,...,|\mathcal{B}|$, we draw i.i.d. parameter samples $\vect{w}^{(i)}=\vmu + \exp(\vect{\psi})\vect{\epsilon}^{(i)}$ with $\vect{\epsilon}^{(i)} \sim \mathcal{N}(\vect{0}, \mathbf{I})$ for \emph{MFG}, or $\vect{w}^{(i)}=\vl^{(c_i)}\vr^{(c_i)}\circ\bar{\vw}$ with $c_i \sim \text{Uniform}\{1,...,C\}$ for \emph{PSE}.
% We then approximate the \emph{expected log-likelihood} by
% \vspace{-0.15cm}
\begin{equation}
% \vspace{-0.35cm}
\small
\label{eq:loss2}
    \mathcal{L}_{ell}^* =  \frac{1}{|\mathcal{B}|}\sum_{i=1}^{|\mathcal{B}|} \log p(y^{(i)}|\vx^{(i)};\vect{w}^{(i)}),\; \vect{w}^{(i)}\stackrel{\text{i.i.d.}}{\sim} q(\vw|\vtheta), \;i = 1,...,|\mathcal{B}|.
\end{equation}

\begin{figure}
\centering
% \begin{subfigure}[b]{.5\textwidth}
%   \centering
  \includegraphics[width=0.5\linewidth]{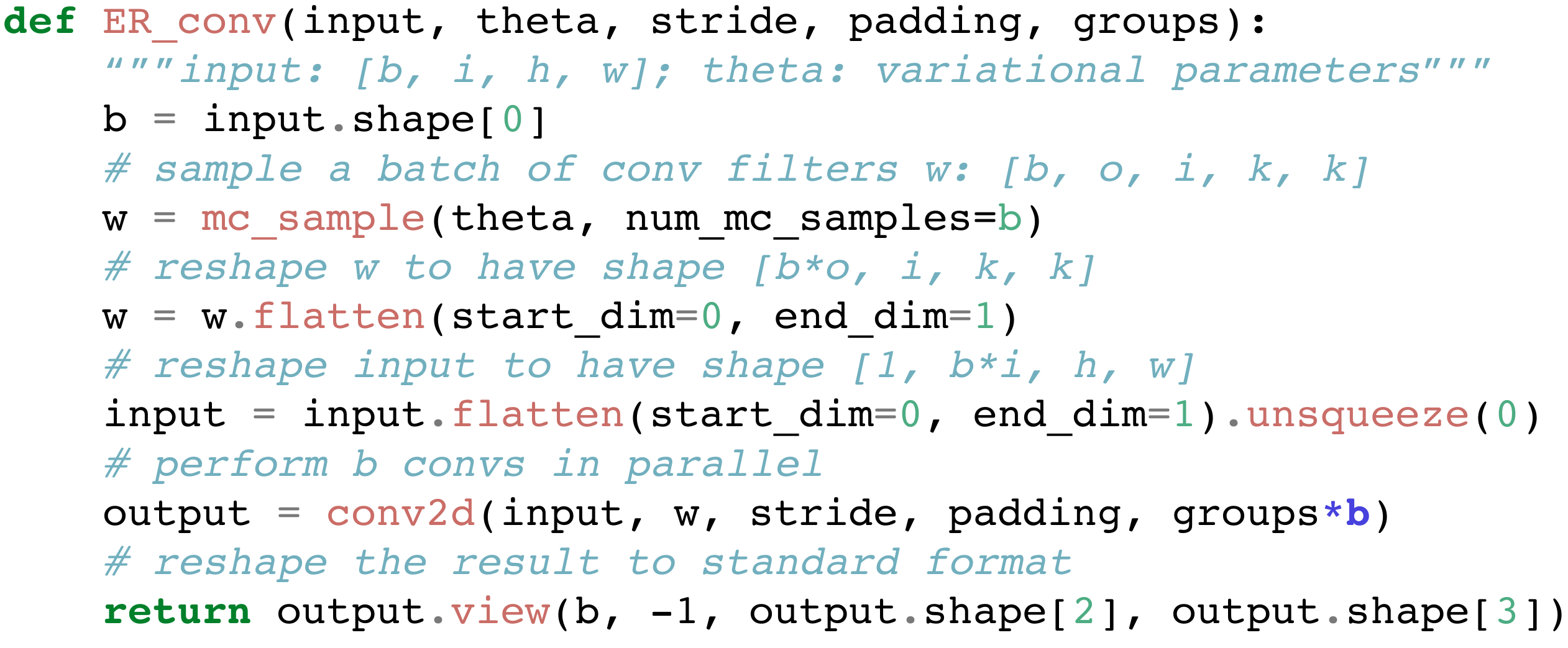}
%   \captionof{figure}{\small Implementation of \emph{exemplar reparametrization} for 2D convolution in PyTorch~\citep{paszke2019pytorch} style.}
%   \label{fig:ewconv}
% \end{subfigure}%
% \hfill
% \begin{subfigure}[b]{.45\textwidth}
%   \centering
  \includegraphics[width=0.45\linewidth]{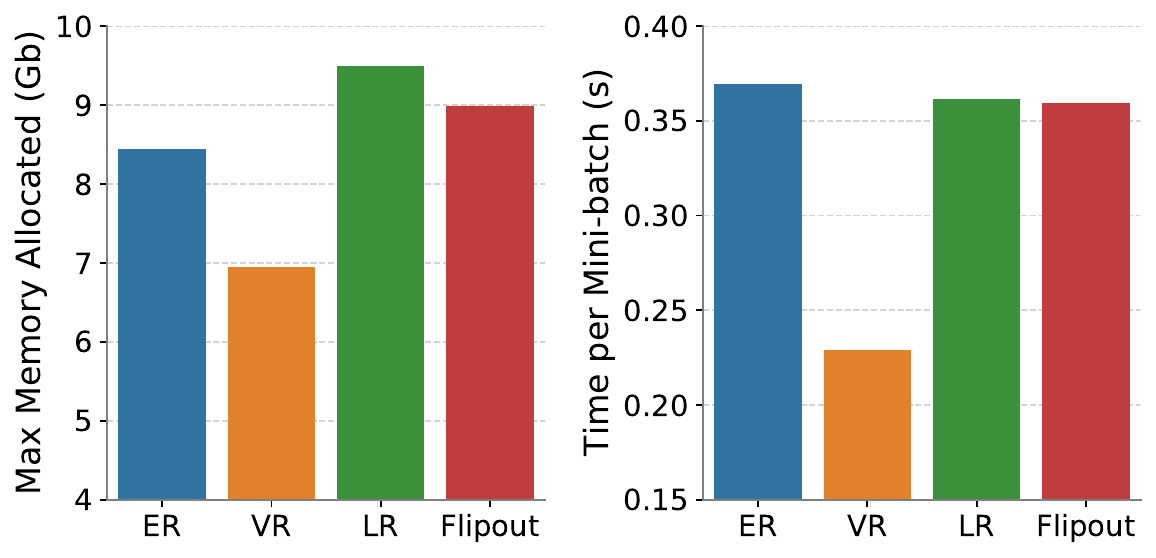}
%   \captionof{figure}{\small Memory and time cost comparison on ImageNet with ResNet-50 architecture. Batch size is 32.}
%   \label{fig:reparam-cost}
% \end{subfigure}
% \vspace{-.3ex}
\caption{(Left): Implementation of \emph{exemplar reparametrization} for 2D convolution in PyTorch~\citep{paszke2019pytorch}. (Right): Memory and time cost comparison among \emph{exemplar reparametrization} (ER), {vanilla reparametrization} (VR)~\citep{kingma2013auto}, {local reparametrization} (LR)~\citep{kingma2015variational}, and Flipout~\citep{wen2018flipout} with mean-field Gaussian variational used (estimated on ImageNet with ResNet-50 architecture).}
\label{fig:ewconv}
% \vspace{-2ex}
\end{figure}

We can see that the computational cost of ER is identical to that of vanilla reparameterization, but ER was criticized for that the involved exemplar-wise computations could not be efficiently done within the popular computation libraries in 2015~\citep{kingma2015variational}.
% It is pleasant to notice that though ER generates more parameters at training, they are temporary, and the resultant computational FLOPS are
%Though orders of magnitude more parameters are generated temporarily during training, the FLOPS is
% identical to that of vanilla reparameterization.
With the rapid development of high-performance device-propriety kernel backends (e.g. cuDNN~\citep{chetlur2014cudnn}) in recent years, we wonder \emph{is the criticism still hold}?
% challenge this criticism and wonder if ER could be accelerated nowadays.
% The challenge of ER is to cope with nowadays ML frameworks and maintain computing efficiency, because off-the-shelf computation kernels typically proceed by sharing parameters among the data in a mini-batch.
% Applying this approach requires converting existing operators in autodiff libraries to be instance-wise, which is not readily available, while still maintaining efficiency.
To this end, we first refurbish ER to fit nowadays ML frameworks.
Our key insight here is to perform multiple exemplar-wise computations in parallel with a single kernel launch, e.g., organize exemplar-wise matrix multiplications as a \texttt{batch matrix multiplication}; organize exemplar-wise convolutions as a \texttt{group convolution} (see Figure~\ref{fig:ewconv} (Left)).
We then conduct an empirical study on the computation cost of ER and relevant methods using \emph{MFG} variational.
Figure~\ref{fig:ewconv} (Right) shows the results.
% We present an example in Figure~\ref{fig:ewconv} on how the standard \texttt{convolution} op can be converted into its exemplar version without compromising computational efficiency. The key insight here is that multiple exemplar convolutions can be expressed as a group convolution, which can be performed in parallel using a single group convolution kernel, leveraging the optimized implementations provided by various device-propriety kernel backends (e.g. cuDNN~\citep{chetlur2014cudnn}).
% Other common operators such as matrix multiplication are straightforward to handle (refer to Appendix~A).

Surprisingly, ER's time efficiency is comparable with that of local reparameterization~\citep{kingma2015variational} and Flipout~\citep{wen2018flipout}, while its memory cost is even lower.
This is perhaps because local reparameterization and Flipout both need to calculate and store one extra mini-batch of feature maps, which are rather large in ImageNet models.
Note that the added memory cost of ER upon vanillar reparameterization comes from the storage of a mini-batch of temporary parameters.
% Their sizes both increase monotonically with thebatch size. The training time of ER is slightly longer thanthe baselines despite withless FLOPS
We clarify that the primary merit of ER over existing methods is the higher generality rather than better learning outcomes.
We hope that ER will benefit the further development of new variational distributions.

\subsection{A Plug-and-play Library}
We wrap the details of the aforementioned modularized stochastic variational inference and ER strategy for \emph{MFG} and \emph{PSE} in a plug-and-play Python library %(will be released after acceptance)
to free the users from the difficulties of implementing \emph{BayesAdapter}.

\section{Experiments}
% \vspace{-0.5ex}
\label{sec:exp}
We apply \emph{BayesAdapter} to a diverse set of benchmarks for empirical verification. % of \emph{BayesAdapter}.

% \hao{add a paragraph summarize the questions answered in experiments.}
% \noindent
\textbf{Settings.} In general, we pre-train DNNs following standard protocols or fetch the pre-trained checkpoints available online, and then perform \emph{Bayesian fine-tuning}. %using the two developed variationals.
We randomly initialize the newly added variational parameters (e.g., $\vect{\psi}$, $\vl^{(c)},\vr^{(c)}$).
Unless otherwise stated, we set $r=1$ and $C=20$ for \emph{PSE} and use the ER trick during training.
We use $S=20$ MC samples to make prediction and quantify \emph{epistemic} uncertainty.
% For \emph{BayesAdapter w/ reg}, we set $\alpha=3$ without tuning and set uncertainty threshold $\gamma=0.75$ according to an observation that the normal examples usually present $<0.75$ mutual information uncertainty across various scenarios.
We conduct experiments on 8 RTX 2080Ti GPUs.
Full details are deferred to Appendix~B. % whenever the space is enough.

\begin{table*}[t]
% \vspace{-1ex}
  \caption{\small Comparison on test accuracy and negative log-likelihood (NLL) on CIFAR-10 benchmark. We use \underline{underline} to emphasize the results obtained given significantly more training effort. For \emph{BayesAdapter}, we repeat every experiment for 3 times and report the error bar.}
%   \vspace{-0.5ex}
  \centering
 \small
  \begin{tabular}{
    >{\raggedright\arraybackslash}p{36ex}%
    >{\raggedleft\arraybackslash}p{28ex}%
    >{\raggedleft\arraybackslash}p{28ex}%
    }%|p{17.5ex}<{\centering}}
  \toprule
% \multirow{1}{*}{Method}& \multicolumn{2}{c}{CIFAR-10} \\
% \cline{2-5}
Method & Accuracy (\%) $\uparrow$ & NLL $\downarrow$ \\
\midrule
\emph{MAP} & 96.92 & 0.1312\\
\emph{Laplace Approx.} & 96.41 & 0.1204\\
\emph{MC Dropout}& 96.95 & 0.1151 \\
\emph{SWAG}& 96.32 & 0.1122 \\
\emph{Deep Ensemble}& \textbf{\underline{97.40}} & \textbf{\underline{0.0869}} \\
\emph{VBNN (MFG)} & 96.95 & 0.0994\\
\emph{VBNN (PSE)} & 96.88 & 0.1328\\
% \hline
\emph{BayesAdapter (MFG)} & 97.10$\pm$0.03 & 0.1007$\pm$0.0014\\
\emph{BayesAdapter (PSE)} & \textbf{97.13}$\pm$0.03 & \textbf{0.0936}$\pm$0.0010\\
% \hline
% \emph{BayesAdapter w/ reg (MFG)} & 96.82$\pm$0.07 & 0.1004$\pm$0.0026 & \textbf{0.985}$\pm$0.005 & \textbf{0.996}$\pm$0.002\\
% \emph{BayesAdapter w/ reg (PSE)} & 96.86$\pm$0.06 & 0.1173$\pm$0.0030 & 0.783$\pm$0.054 & \textbf{0.998}$\pm$0.001\\
% \emph{BayesAdapter} & 96.82$\pm$0.07 & 0.1004$\pm$0.0026\\
  \bottomrule
   \end{tabular}
  \label{table:clf-cifar}
%   \vspace{-3ex}
% \vspace{-0.2cm}
\end{table*}

\textbf{Baselines.} We consider extensive baselines including:
(1) \emph{MAP}, which is the fine-tuning start point, %\footnote{The converged parameters of \emph{MAP} are used as the fine-tuning start of \emph{BayesAdapter-} and \emph{BayesAdapter}.},
(2) \emph{Laplace Approx.}: which preforms Laplace approximation with diagonal Fisher information matrix,
(3) \emph{MC Dropout}, which is a dropout variant of \emph{MAP},
(4) \emph{VBNN}, which refers to from-scratch trained variational BNNs.
In particular, the variational BNN methods like \emph{BayesAdapter} and \emph{VBNN} are evaluated on both the \emph{MFG} and \emph{PSE} variationals.
We also include \emph{Deep Ensemble}~\citep{lakshminarayanan2017simple}, and \emph{SWAG}~\citep{maddox2019simple} %, whose performance is not worse than SGLD~\citep{welling2011bayesian},  KFAC Laplace~\citep{mackay1992practical} and temperature scaling~\citep{guo2017calibration},
as baselines on CIFAR-10 benchmark~\citep{krizhevsky2009learning}.\footnote{Currently, we have not scaled \emph{Deep Ensemble} and \emph{SWAG}, whihc both require storing tens of NN weights copies, up to ImageNet due to resource constraints.}
% We take the results of \emph{SWAG} from its paper due to implementation difficulties.

% \noindent \textbf{Metrics.} We concern  %to estimate the predictive performance and the uncertainty estimation of the trained models, respectively:
% $(\RN{1})$ the \emph{posterior predictive} performance; $(\RN{2})$ the average precision (AP) of directly using the mutual information uncertainty (Eq.~(\ref{eq:mi})) to distinguish OOD test samples (labeled 1) from normal test samples (labeled 0).
% Eq.~(\ref{eq:mi}) of the deterministic baseline \emph{MAP} is 0, so we take the predictive entropy as an alternative uncertainty measure for \emph{MAP}.

% \vspace{-1.5ex}
\subsection{CIFAR-10 Classification}
% \vspace{-.5ex}
\label{sec:perf}

We first conduct experiments on CIFAR-10 with wide-ResNet-28-10 architecture~\citep{zagoruyko2016wide}.
We perform \emph{Bayesian fine-tuning} for 12 epochs with the weight decay coefficient $\lambda$ set as 2e-4.
Table~\ref{table:clf-cifar} outlines the comparison on prediction performance.

It is worth noting that \emph{BayesAdapter} substantially outperforms \emph{MAP}, \emph{Laplace Approx.}, and \emph{MC Dropout} in aspect of predictive performance. %, highlighting the practical value of our workflow.
\emph{BayesAdapter} also surpasses \emph{SWAG} due to that \emph{SWAG} does not directly benefit from high-performing pre-trained MAP models.
The accuracy upper bound is \emph{Deep Ensemble}, which trains 5 isolated \emph{MAP}s and assembles their predictions to explicitly investigate diverse function modes, but it is much more expensive than \emph{BayesAdapter}.
\emph{VBNN} is clearly defeated by \emph{BayesAdapter}, confirming our claim that performing \emph{Bayesian fine-tuning} from the converged deterministic checkpoints is beneficial to explore more qualified posteriors.
\emph{BayesAdapter (PSE)} surpasses \emph{BayesAdapter (MFG)}, especially in the aspect of NLL.
Unless specified otherwise, we refer to \emph{BayesAdapter (PSE)} as \emph{BayesAdapter} in the following.

\textbf{Converged ELBO.} We compare the converged (training) ELBO of  \emph{BayesAdapter} and \emph{VBNN}:
the former gives $\mathcal{L}_{ell}=-0.019$ and $\mathcal{L}_c=-2806.8$ while the latter gives  $\mathcal{L}_{ell}=-0.032$ and $\mathcal{L}_c=-2384.3$.
% The KL term ($\mathcal{L}_c$) is \emph{huge} as the architecture has more than 30M parameters.
This implies that \emph{Bayesian fine-tuning} makes the approximate posterior converge to somewhere with better data fitting than from-scratch VI. %, while the convergence is \emph{distinct} from that of from-scratch variational inference.

\textbf{CIFAR-10 corruptions.}
We then assess the quality of predictive uncertainty on CIFAR-10 corruptions~\citep{hendrycks2019benchmarking}.
Figure~\ref{fig:acc-comp} (Right) shows the results, which reflect the efficacy of \emph{BayesAdapter} for promoting model calibration.

\begin{wrapfigure}{r}{0.38\linewidth}
% \centering\vspace{-2.8ex}
\includegraphics[width=0.98\linewidth]{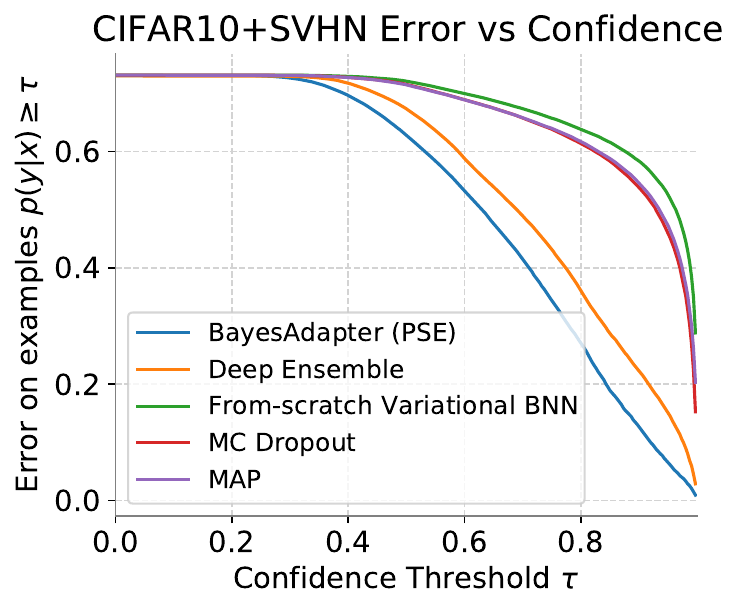}
\vspace{-1.ex}
\caption{\small Error vs. confidence plots for models trained on CIFAR-10 and tested on both CIFAR-10 and SVHN.}% Arguably, they involve roughly the same computations.}
\vspace{-2ex}
\label{fig:svhn}
\end{wrapfigure}
% As can be seen from the discussion so far, we only supervise the BNN to fit the data, leaving the uncertainty quantification solely originating from Bayes' principle.
% Based on the efficient \emph{BayesAdapter} paradigm, we can perform faster learning of variational BNNs, and hence deeper investigations on their properties.
\textbf{CIFAR-10 vs SVHN.} %We then compare the predictive uncertainty of \emph{BayesAdapter} as well as the baselines under regular domain shift.
Following \cite{he2020bayesian}, we evaluate the trained models on both CIFAR-10 and SVHN~\citep{netzer2011reading} test sets.
% We experiment with PSE due to its better performance than MFG (shown above).
For every confidence threshold $0\leq\tau<1$, we compute the average error rate for predictions with $\geq \tau$ confidence (all predictions on SVHN data are regarded as incorrect).
We depict the error vs. confidence curves in Figure~\ref{fig:svhn}.
It is clear that \emph{BayesAdapter (PSE)} has made more conservative predictions on the out-of-distribution (OOD) SVHN data than all baselines.
% We perform only \textbf{12} epochs of \emph{Bayesian fine-tuning} upon MAP, while the model calibration is significantly promoted.
\emph{BayesAdapter (PSE)} even outperforms the expensive Deep Ensemble, implying that the parameter-sharing mechanism may impose further regularization on learning.
The comparison also confirms from-scratch variational BNNs, even with the PSE variational, have difficulties to find good posteriors.

% \noindent
\textbf{Speedup.}
% To interpret the speedup of \emph{BayesAdapter} over \emph{VBNN}, w
% Based on our implementation, SVI with \emph{PSE} takes around $2$ minutes for one epoch.
BayesAdapter requires 200 epochs of deterministic training plus 12 epochs of variational training, while VBNN requires 200 epochs of variational training. Considering the cost of variational training is several times (about 2.1$\times$) that of deterministic training, the training time saved by BayesAdapter is considerable.
% Thus, \emph{VBNN} trained from scratch consumes $400$ minutes for 200-epoch training, while \emph{BayesAdapter} needs only $24$ minutes for 12-epoch fine-tuning, saving $376$ minutes ($\bf 94\%$) training time than \emph{VBNN}.

% take a look at the training time of them on ImageNet with ResNet-50 using 8 RTX 2080Ti GPUs,
% Practically, the training of the DNN counterpart of \emph{BayesAdapter} (i.e., \emph{MAP}) consumes 15 hours for 90 epochs by taking automatic mixed precision technique.
% The \emph{Bayesian fine-tuning} for

\begin{table*}[t]
% \vspace{-1ex}
  \caption{\small Comparison on test accuracy and NLL on ImageNet benchmark.}
%   \vspace{-0.5ex}
  \centering
 \small
  \begin{tabular}{
    >{\raggedright\arraybackslash}p{36ex}%
    >{\raggedleft\arraybackslash}p{28ex}%
    >{\raggedleft\arraybackslash}p{28ex}%
    }%|p{17.5ex}<{\centering}}
  \toprule
% \multirow{2}{*}{Method}& \multicolumn{2}{c||}{CIFAR-10} & \multicolumn{2}{c}{ImageNet} \\
% \cline{2-5}
Method & Accuracy (\%) $\uparrow$ & NLL $\downarrow$ \\
\midrule
\emph{MAP} & 76.13 &0.9618 \\
\emph{Laplace Approx.} & 75.89 & 0.9739\\
\emph{MC Dropout}& 74.88 &0.9884\\
% \emph{SWAG}& 96.32 & 0.1122 & - & - \\
% \emph{Deep Ensemble}& \textbf{\underline{97.40}} & \textbf{\underline{0.0869}} & - & -\\
\emph{VBNN (MFG)} & 75.97 & 0.9435\\
\emph{VBNN (PSE)} & 75.12 & 0.9865\\
% \hline
\emph{BayesAdapter (MFG)} & {76.45}$\pm$0.05  & {0.9303}$\pm$0.0005 \\
\emph{BayesAdapter (PSE)} & \textbf{76.80}$\pm$0.03  & \textbf{0.9159}$\pm$0.0010\\
% \hline
% \emph{BayesAdapter w/ reg (MFG)} & 96.82$\pm$0.07 & 0.1004$\pm$0.0026 & \textbf{0.985}$\pm$0.005 & \textbf{0.996}$\pm$0.002\\
% \emph{BayesAdapter w/ reg (PSE)} & 96.86$\pm$0.06 & 0.1173$\pm$0.0030 & 0.783$\pm$0.054 & \textbf{0.998}$\pm$0.001\\
% \emph{BayesAdapter} & 96.82$\pm$0.07 & 0.1004$\pm$0.0026\\
  \bottomrule
   \end{tabular}
  \label{table:clf-in}
%   \vspace{-3ex}
% \vspace{-0.2cm}
\end{table*}

% \vspace{-1.2ex}
\subsection{ImageNet Classification}
% \vspace{-.2ex}
We then scale up \emph{BayesAdapter} to ImageNet with ResNet-50~\citep{he2016deep} architecture.
We launch fine-tuning for merely \textbf{4 epochs} with the weight decay coefficient $\lambda$ set as 1e-4.

Table~\ref{table:clf-in} reports the empirical comparison.
As expected, most results are consistent with those on CIFAR-10.
On this large-scale scenario, it is more clear that the from-scratch learning baseline \emph{VBNN} would suffer from local optima.
The striking improvement of \emph{BayesAdapter} upon \emph{MAP} validates the benefits of Bayesian treatment.
Zooming in, we also note that \emph{BayesAdapter (PSE)} reveals remarkably higher accuracy than \emph{BayesAdapter (MFG)}, testifying the superior expressiveness of \emph{PSE} over \emph{MFG}.
% On the other side of the spectrum, \emph{BayesAdapter w/ reg} is better than its fine-tuning start point \emph{MAP} and the from-scratch baseline \emph{VBNN} on both axes of comparison, double evidencing the usefulness of the uncertainty regularization and highlighting the practical value of the proposed approach.

\begin{table*}[t]
% \vspace{-0.5ex}
  \caption{\small Accuracy $\uparrow$ comparison on open-set face recognition with MobileNetV2 architecture.}
%   \vspace{-0.25cm}
  \centering
 \small
  \begin{tabular}{>{\raggedright\arraybackslash}p{28ex}%
    >{\raggedleft\arraybackslash}p{11ex}%
    >{\raggedleft\arraybackslash}p{11ex}%
    >{\raggedleft\arraybackslash}p{11ex}%
    >{\raggedleft\arraybackslash}p{11ex}%
    >{\raggedleft\arraybackslash}p{11ex}%
    } %|p{10ex}<{\centering}|p{11ex}<{\centering}}
  \toprule
{Method}& LFW  & CPLFW & CALFW & CFP-FF & CFP-FP\\ % & VGGFace2 & AgeDB-30\\
\midrule
\emph{MAP} & 98.2\% & 84.0\% & {87.6\%} & \textbf{97.8}\% & 92.7\%\\ % & 91.7\% & 85.3\% \\
\emph{MC Dropout}& 98.2\% & 83.6\% & 87.3\% & \textbf{97.8}\%& 92.8\% \\% & {92.6\%} & \textbf{86.0\%}\\
% \emph{VBNN (MFG)} & 97.8\% & 82.4\% & 85.7\% & 96.8\% & 91.4\% & 90.5\% & 83.8\%\\
% \hline
\emph{BayesAdapter (MFG)} & \textbf{98.4}\% & {83.9\%} & 85.8\% & 97.6\% & 92.9\%\\% & \textbf{93.1}\% & 84.5\% \\
\emph{BayesAdapter (PSE)} & \textbf{98.4}\% & \textbf{84.7\%} & \textbf{87.8}\% & \textbf{97.8}\% & \textbf{93.1}\%\\% & 92.4\% & 85.7\% \\
% \hline
% \emph{BayesAdapter w/ reg (MFG)} & \textbf{98.4\%} & 84.1\% & 87.4\% & \textbf{97.9\%} & \textbf{93.1\%}\\% & {92.5\%} &  84.8\%\\
% \emph{BayesAdapter w/ reg (PSE)} & {98.0\%} & 84.1\% & 87.7\% & \textbf{97.9\%} & {92.6\%}\\% & {92.0\%} &  85.9\% \\
  \bottomrule
   \end{tabular}
  \label{table:face}
%   \vspace{-2.ex}
% \vspace{-0.2cm}
\end{table*}

\begin{table}[t]
% \vspace{-2ex}
\small
  \caption{\small Ablation study on the rank $r$ of \emph{PSE}. (ImageNet)}
%   \vspace{-0.25cm}
  \centering \small
  %\begin{sc}
  \begin{tabular}{>{\raggedright\arraybackslash}p{36ex}%
    >{\raggedleft\arraybackslash}p{28ex}%
    >{\raggedleft\arraybackslash}p{28ex}%
  }
  \toprule
  Method & Accuracy (\%) & \# of Param. (M)\\
  \midrule
  \emph{MAP} & 76.13 & 25.56 \\
  \emph{BayesAdapter (PSE, $r$=1)} & 76.80 & 27.21 \\
  \emph{BayesAdapter (PSE, $r$=8)} & 76.78 & 38.76\\
  \emph{BayesAdapter (PSE, $r$=16)} & 76.80 & 51.95 \\
  \bottomrule
   \end{tabular}
 % \end{sc}
%  \vspace{-0.3cm}
  \label{table:abl-r}
%   \vspace{-1ex}
\end{table}

% \vspace{-1.2ex}
\subsection{Face Recognition}
% \vspace{-.2ex}
To demonstrate the universality of \emph{BayesAdapter}, we further apply it to the challenging face recognition task based on MobileNetV2 architecture~\citep{sandler2018mobilenetv2}.
We train models on the CASIA dataset~\citep{yi2014learning}, and perform comprehensive evaluation on face verification datasets including LFW~\citep{huang2008labeled}, CPLFW~\citep{zheng2018cross}, CALFW~\citep{zheng2017cross}, and CFP~\citep{sengupta2016frontal}. %, VGGFace2~\citep{cao2018vggface2}, and AgeDB-30~\citep{moschoglou2017agedb}. %, following standard protocols.
We launch fine-tuning for 4 epochs with $\lambda=5e-4$.
% Given the ineffectiveness of the from-scratch learning of variational BNNs and Laplace approximation reflected by the above studies,
We compare our method to \emph{MAP} and \emph{MC Dropout}, two popular baselines in face recognition.
We depict the recognition accuracy in Table~\ref{table:face}. % and that on uncertainty estimates in Table 1 and 2 of Appendix C.

It is noteworthy that Bayesian principle can induce better predictive performance for face recognition models.
\emph{BayesAdapter (PSE)} has outperformed the fine-tuning start point \emph{MAP} and the popular baseline \emph{MC Dropout} in most verification datasets, despite being fine-tuned for only several rounds.
% The near-perfect results of \emph{BayesAdapter w/ reg} for detecting OOD instances signify the potential of the developed approach in industrial applications.

% In addition, we notice that the uncertainty estimates of \emph{BayesAdapter w/ reg (PSE)} for detecting OOD data are relatively less calibrated than \emph{BayesAdapter w/ reg (MFG)}.
% We deduce that this is because once \emph{MFG} approaches a flat high-quality posterior, \emph{MFG} can dedicate its variance parameters to accounting for the uncertainty regularization, while \emph{PSE} needs to find diverse posterior spikes to guarantee both data fittingness and calibrated uncertainty concurrently, which may be more demanding.
% The calibrated uncertainty estimation can help to realise more robust decision making by refusing to predict for unfamiliar data.

% \vspace{-1.5ex}
\subsection{More Empirical Analyses}
% \vspace{-.5ex}
\label{sec:abl}

\begin{wraptable}{r}{0.55\linewidth}
\vspace{-1.ex}
\small
  \caption{\small Comparison on model calibration (ECE $\downarrow$). %We omit the face recognition tasks where features are leveraged for classification directly.
  }
  \vspace{-1ex}
  \centering \footnotesize
  %\begin{sc}
  \begin{tabular}{>{\raggedright\arraybackslash}p{24ex}%
    >{\raggedleft\arraybackslash}p{12ex}%
    >{\raggedleft\arraybackslash}p{12ex}%
  }
  \hline
  Method & CIFAR-10 & ImageNet\\
  \hline
\emph{MAP} & 0.0198 & 0.0373\\
% \emph{Laplace Approx.} & 0.0106 & 0.0375\\
% \emph{MC Dropout}& 0.0119 & {0.0152}\\
\emph{SWAG}& 0.0088 & -\\
\emph{Deep Ensemble}& \underline{\textbf{0.0057}} & -\\
\emph{VBNN (MFG)} & {0.0074} & 0.0183\\
\emph{VBNN (PSE)} & 0.0188 & 0.0202\\
\hline
\emph{BayesAdapter (MFG)} & 0.0091 & 0.0289\\
\emph{BayesAdapter (PSE)} & \textbf{0.0058} & \textbf{0.0129}\\
% \hline
% \emph{BayesAdapter w/ reg (MFG)} & \textbf{0.0057} & 0.0165\\
% \emph{BayesAdapter w/ reg (PSE)} & 0.0136 & 0.0521\\
  \hline
   \end{tabular}
 % \end{sc}
%  \vspace{-0.4cm}
  \label{table:ece}
  \vspace{-2.2ex}
\end{wraptable}
\textbf{Model calibration on in-distribution data.}
We estimate the model calibration, measured by ECE, of various methods on in-distribution data, and report the results in Table~\ref{table:ece}.
The ECE of \emph{SWAG} on ImageNet is based on ResNet-152~\citep{he2016identity}.
Notably, the ECE of \emph{BayesAdapter (PSE)} is on par with \emph{Deep Ensemble}, significantly better than the other baselines.

\textbf{The impact of the rank $r$ for \emph{PSE}.}
As stated, we set $r=1$ for all the above studies for maximal parameter saving.
Yet, does the small rank $r$ confine the expressiveness of \emph{PSE}?
We perform an ablation study to pursue the answer.
As shown in Table~\ref{table:abl-r}, the capacity of \emph{PSE} can already be sufficiently unleashed when the rank is 1, where only marginally added parameters are introduced over MAP.
This indicates the merit of \emph{PSE} for efficient learning.

\textbf{The effectiveness of {exemplar reparameterization}.}
We build a toy model with only a convolutional layer and fix the model input and the target output.
We employ the \emph{MFG} variational on the convolutional parameters and computing the variance of stochastic gradients across 500 runs.
We average the gradient variance of $\vmu$ and $\vect{\psi}$ over all their coordinates, and observe that vanilla reparameterization typically introduces $100\times$ more variance than \emph{exemplar reparameterization}. %, despite with the same FLOPS.
% We provide a visualization of the stochastic gradients of $\vmu$ for the two convolutions in Figure~\ref{fig:grad-var} in Appendix~\ref{app:gradvar}.
% \zhijie{add lr into comparison}
% 8.201443e-09 2.7192462e-10
% 1.0598767e-06 3.4679104e-08

\begin{figure}[t]
\small
% \vspace{-1ex}
     \centering
    %  \begin{subfigure}[b]{0.45\textwidth}
        %  \centering
         \includegraphics[width=0.45\linewidth]{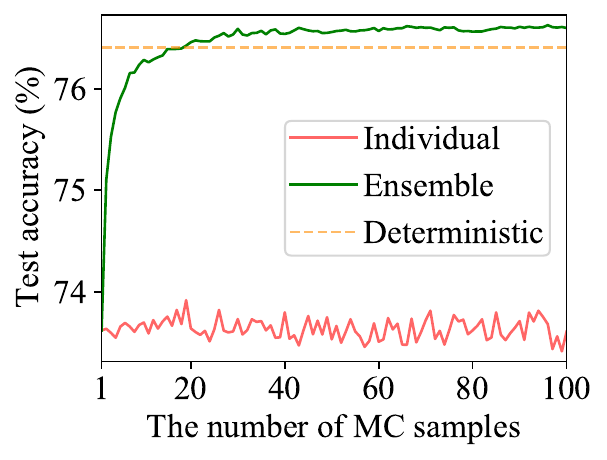}
        % \vspace{-1.5ex}
        % \caption{}
        % \label{fig:mode_col}
    %  \end{subfigure}
    %  \hfill
    %  \begin{subfigure}[b]{0.45\textwidth}
        %  \centering
         \includegraphics[width=0.45\linewidth]{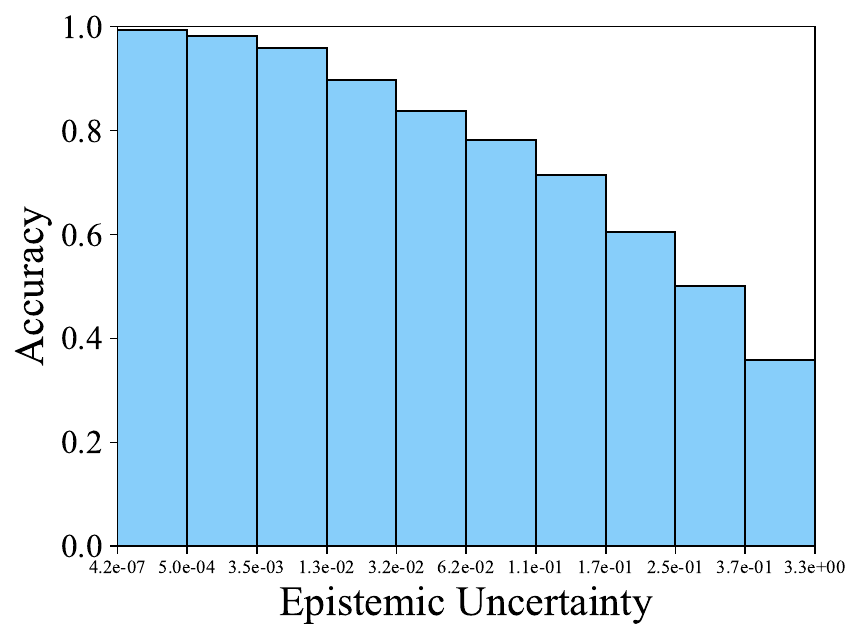}
        % \vspace{-1.5ex}
        % \caption{}%As shown, the accuracy drops as the uncertainty increases.}
        % \label{fig:rej-dec}
    %  \end{subfigure}
     \vspace{-2ex}
     \caption{\small (Left): \small Test accuracy varies w.r.t. the number of MC samples for \emph{Bayes ensemble}.
     (Right): Comparison on the accuracy for instance buckets of equal size but with rising uncertainty. (\emph{BayesAdapter (MFG)}, ImageNet)}
     \label{fig:abl}
     \vspace{-2ex}
\end{figure}

\textbf{The impact of ensemble number.}
% gpu31: /data/zhijie/snapshots\_ab/ft-gan1000-.75
% gpu30: /data/zhijie/snapshots\_ab\_in/ft-gan-.75-1
We draw the change of test accuracy w.r.t. the number of MC samples $S$ for \emph{Bayes ensemble} in Figure~\ref{fig:abl} (Left).
The model is trained by \emph{BayesAdapter (MFG)} on ImageNet.
The points on the red line represent the individual accuracies of the 100 parameter samples.
The yellow dashed line refers to the deterministic inference with only the Gaussian mean.
The green line displays the effects of \emph{Bayes ensemble} -- the predictive performance increases from $<74\%$ to $>76\%$ quickly before seeing 20 parameter samples, and gradually saturates after that.
% That is why we set $S=20$ in the above studies.

\textbf{Uncertainty-based rejective decision.}
In practice, we expect our models to be accurate on the data that they are certain about.
In this spirit, we gather the \emph{epistemic} uncertainty estimates for ImageNet validation data given by \emph{BayesAdapter (MFG)}, based on which we divide the data into 10 buckets of equal size but with increasing uncertainty.
We depict the average accuracy of each bucket in Figure~\ref{fig:abl} (Right).
As expected, our BNN is more accurate for instances with smaller uncertainty.
% Quantitatively, there are 95\% instances with uncertainty less than 0.45, and their accuracy is 78.6\%;
% there are 90\% instances with uncertainty less than 0.37, and their accuracy is 80.7\%;
% there are 80\% instances with uncertainty less than 0.25, and their accuracy is 84.8\%.

% \vspace{-1.5ex}
\section{Related Work}
% \vspace{-.5ex}
Fruitful works have emerged in the BNN community in the last decade~\citep{graves2011practical,welling2011bayesian,blundell2015weight,kingma2013auto,balan2015bayesian,liu2016stein,kendall2017uncertainties,wu2018deterministic}.
% A particularly appealing line is to develop stochastic variational inference methods for BNNs to simplify the inference problem as an optimization one~\citep{blundell2015weight,kingma2013auto}.
However, most of the existing works cannot achieve the goal of practicability.
For example, some works trade learning efficiency for flexible variational posteriors, leading to restrictive scalability~\citep{louizos2016structured,louizos2017multiplicative,shi2018kernel,sun2018functional}.
\cite{khan2018fast,zhang2018noisy,osawa2019practical} build Adam-like optimizers to do variational inference, but their parallel training throughput and compatibility with data augmentation are inferior to SGD.
Approximate Bayesian methods like Monte Carlo dropout~\citep{gal2016dropout} and Deep Ensemble~\citep{lakshminarayanan2017simple} can maintain good predictive performance but suffer from degenerated uncertainty estimates~\citep{fort2019deep} or high cost.
% What's worse, the existing works usually evaluate on impractical OOD data~\citep{louizos2017multiplicative,pawlowski2017implicit} to show the merit of Bayesian principle.
% Instead, we offer a new evaluation standard in this work, which may benefit the following works.

Laplace approximation~\citep{mackay1992practical,ritter2018scalable} is a known approach to transform a DNN to a BNN, but it is inflexible due to its postprocessing nature and some strong assumptions made for practical concerns.
% inefficient due to the manipulation of Hessian.
Alternatively, \emph{BayesAdapter} works in the style of fine-tuning, which is more natural and economical for deep networks.
Bayesian modeling the last layer of a DNN is proposed recently~\citep{kristiadi2020being}, and its combination with \emph{BayesAdapter} deserves an investigation.
\emph{BayesAdapter} connects to MOPED~\citep{krishnan2020specifying} in that their variational configurations are both based on \emph{MAP}.
% With the prior specified as MAP mean and unit variance, the primary objectives of MOPED are also to speed up the learning and to bypass the potential local optimas of the posterior.
However, MOPED solves the prior specification problem for BNNs while \emph{BayesAdapter} constitutes a practical framework to bring variational BNNs to the masses.
In detail, MOPED uses MAP to define the prior while \emph{BayesAdapter} uses MAP to initialize the parameters of the variational distribution. Beyond this, this work makes valuable technical contributions including the \emph{PSE} variational and the refinement of \emph{exemplar reparameterization}.
We have also done a thorough study on how pre-training benefits VI.
% Anyway, we provide an empirical comparison between \emph{BayesAdapter} and MOPED in Appendix C.
Moreover, the results in Appendix C show that MOPED suffers from more serious over-fitting than BayesAdapter.

% \emph{BayesAdapter}
% is further designed to achieve good user-friendliness, improved learning stability, and trustable uncertainty estimation, which are essentially crucial in real-world and large-scale settings.

\vspace{-.5ex}
\section{Conclusion}
% \vspace{-.5ex}
This work proposes the \emph{BayesAdapter} framework to ease the learning of variational BNNs.
Our core idea is to perform \emph{Bayesian fine-tuning} instead of expensive from-scratch Bayesian learning.
We develop plug-and-play implementations for the stochastic variational inference under two representative variational distributions, % to make the fine-tuning more user-friendly,
and refine \emph{exemplar reparameterization} to efficiently reduce gradient variance. %, making the training more stable.
% We also propose a biased yet meaningful uncertainty regularization to calibrate the \emph{epistemic} uncertainty of variational BNNs.
We evaluate \emph{BayesAdapter} in diverse scenarios and report promising results.
One limitation of \emph{BayesAdapter} is that practitioners may need to carefully tune the optimization configurations for \emph{Bayesian fine-tuning} to achieve reasonable performance.
Regarding future work, the application of BayesAdapter to more exciting scenarios like contextual bandits deserves further investigation.

\vspace{-.5ex}
\section*{Acknowledgments}
% \vspace{-.5ex}
This work was supported by NSFC Projects (Nos. 62061136001, 62076145, 62076147, U19B2034, U1811461, U19A2081, 61972224), Beijing NSF Project (No. JQ19016), BNRist (BNR2022RC01006), Tsinghua Institute for Guo Qiang, and the High Performance Computing Center, Tsinghua University. J.Z is also supported by the XPlorer Prize.

\bibliography{acml22}

% \appendix

% \section{First Appendix}\label{apd:first}

% This is the first appendix.

% \section{Second Appendix}\label{apd:second}

% This is the second appendix.

\end{document}

\maketitle

\appendix

\section{Exemplar Fully-connected Layer}
\label{app:exemplar-op}
As introduced in Sec~3.3, the regular convolution can be elegantly converted into an exemplar version by resorting to group convolution.
The other popular operators are relatively easy to handle.
Take the fully-connected (FC) layer as an example: assuming a feature $x \in \mathbb{R}^{b\times i}$, we draw $b$ i.i.d. FC weights and concatenate them as $w \in \mathbb{R}^{b\times i \times o}$, then invoke \texttt{batch\_matmul}$(x, w)$ to get the output.
% we substitute the qualified batch matrix multiplication, which is highly optimized in the well-known autodiff libraries, for matrix multiplication.
% For the affine transformation in batch normalization~\citep{ioffe2015batch}, we can at first sample dedicated affine weight and bias for every exemplar in the batch, then perform transformation with these two batches of parameters by just not \emph{broadcasting} on the batch dimension.

\section{More Experimental Details}
\vspace{-0.1cm}

\label{app:exp-setting}

The only important hyper-parameter is the weight decay coefficient $\lambda$. Other hyper-parameters for specifying optimization dynamics all follow standard practice in the DL community. 

For $\lambda$, we keep it consistent between pre-training and fine-tuning without elaborated tuning, e.g., $\lambda=2e-4$ for the wide-ResNet-28-10 architecture on CIFAR-10, $\lambda=1e-4$ for ResNet-50 architecture on ImageNet, and $\lambda=5e-4$ for MobileNet-V2 architecture on CASIA. These values correspond to isotropic Gaussian priors with $\sigma_0^2$ as 0.1, 0.0078, and 0.0041 on CIFAR-10, ImageNet, and CASIA, respectively. It is notable that for a ``small'' dataset like CIFAR-10, a flatter prior is preferred. While on larger datasets with stronger data evidence, we need a sharper prior for regularization.

For the pre-training, we follow standard protocols available online.
On CIFAR-10, we perform CutOut~\citep{devries2017improved} transformation upon popular resize/crop/flip transformation for data augmentation.
On ImageNet, we leverage the ResNet-50 checkpoint on PyTorch Hub as the converged deterministic model.
On face tasks, we train MobileNetV2 following popular hyper-parameter settings, and the pre-training takes 90 epochs.
% We use the same weight decay coefficients in both the pre-training and the fine-tuning.

% For the fine-tuning, we set $lr_{\vect{\psi}}$ to follow a cosine schedule decay at 1/4, 1/2, and 3/4 of the total fine-tuning steps from 0.1, and set $lr_{\vmu}$ to be the final value of $lr_{\vect{\psi}}$ on the CIFAR-10, ImageNet, and face recognition benchmarks.
% We add a coefficient 3 before the $\mathcal{L}_{\text{unc}}$ term in Line 8 of Algorithm~\ref{algo:1} for \emph{Bayesian fine-tuning} on ImageNet to achieve better uncertainty calibration.
For models on face recognition, we utilize the features before the last FC layer of the MobileNetV2 architecture to conduct feature distance-based face classification in the validation phase, due to the open-set nature of the validation data.
The \emph{Bayes ensemble} is similarly achieved by assembling features from multiple runs as the final feature for estimating predictive performance.
But we still adopt the output from the last FC layer for uncertainty estimation (i.e., estimating Eq.~(4)).

As for the \emph{MC dropout}, we add dropout-0.3 (0.3 denotes the dropout rate) before the second convolution in the residual blocks in wide-ResNet-28-10, dropout-0.2 after the second and the third convolutions in the bottleneck blocks in ResNet-50, and dropout-0.2 before the last fully connected (FC) layer in MobileNetV2.

For reproducing \emph{Deep Ensemble}, we train 5 \emph{MAP}s separately, and assemble them for prediction and uncertainty quantification.
For reproducing \emph{SWAG}, we take use of its official implementation, and leverage 20 MC samples for prediction.

\section{Comparison Between \emph{BayesAdapter} and MOPED}
We emphasize that MOPED solves the \emph{prior specification} problem for BNNs while \emph{BayesAdapter} constitutes a \emph{practical} framework to \emph{bring variational BNNs to the masses}.
Empirically, we evaluate MOPED on CIFAR-10 with MFG variational, and get 0.0143 training loss ($\mathcal{L}_{ell}$), 96.92\% top1 accuracy, 0.1001 test NLL, and 0.0100 ECE.
Compared to \emph{BayesAdapter}'s results (0.0191, 97.10\%, 0.1007, and 0.0091), we find MOPED exhibits more seriously over-fitting, implying that taking MAP as prior poses under-regularization.

\vspace{-0.1cm}
\section{Visualization of the Learned Posterior}
\begin{figure*}[h!]
% \vspace{-0.5cm}
\centering
% \begin{subfigure}[b]{\textwidth}
% \centering
    \includegraphics[width=0.8\textwidth]{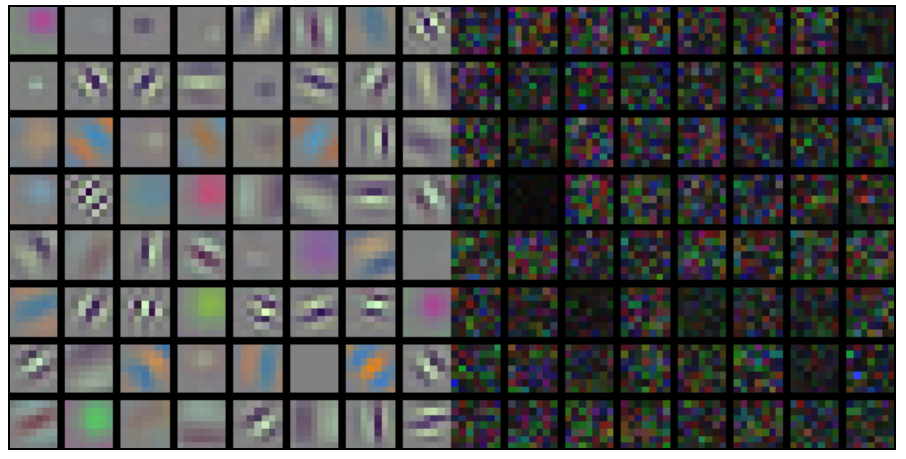}
% \end{subfigure}
\caption{Left: the mean of the \emph{MFG} posterior. Right: the variance of the \emph{MFG} posterior. These correspond to a convolutional kernel with 64 output channels and 3 input channels, where every output channel corresponds to a separate image.}
\label{fig:posterior}
\end{figure*}

\vspace{-0.1cm}
\label{app:visl-firstlayer}
We plot the parameter posterior of the first convolutional kernel in ResNet-50 architecture learned by \emph{BayesAdapter (MFG)} on ImageNet in Figure~\ref{fig:posterior}. 
The learned posterior variance seems to be disordered, unlike the mean. We leave more explanations as future work.

% \section{First Appendix}\label{apd:first}

% This is the first appendix.

% \section{Second Appendix}\label{apd:second}

% This is the second appendix.

%\bibliographystyle{plain}
\bibliography{acml22}